\documentclass{article}

\usepackage[dblblindworkshop, final]{neurips_2025}
\usepackage{graphicx}
\usepackage{amsmath}
\usepackage{amssymb}
\usepackage{algorithm}
\usepackage{algpseudocode}
\usepackage{wrapfig}
\usepackage{caption}

\usepackage[utf8]{inputenc} 
\usepackage[T1]{fontenc}    
\usepackage{hyperref}       
\usepackage{url}            
\usepackage{booktabs}       
\usepackage{amsfonts}       
\usepackage{nicefrac}       
\usepackage{microtype}      
\usepackage{xcolor}         
\workshoptitle{Efficient Reasoning}

\title{ORPO-Distill: Mixed-Policy Preference Optimization for Cross-Architecture LLM Distillation}

\author{%
  Aasheesh Singh\thanks{Corresponding author}\\
  Phi Labs, Quantiphi\\
  Toronto, ON, M5V 2L1, Canada\\
  \texttt{aasheesh.singh@quantiphi.com} \\
  \AND
   Vishal Vaddina\\
  Phi Labs, Quantiphi\\
  Toronto, ON, M5V 2L1, Canada\\
  \texttt{vishal.vaddina@quantiphi.com} \\
\And
  Dagnachew Birru\\
  Phi Labs, Quantiphi\\
  Marlborough, MA, 01752, USA\\
  \texttt{dagnachew.birru@quantiphi.com} \\
}

\begin{document}
\maketitle
\begin{abstract}
We introduce ORPO-Distill, a general-purpose method for cross-architecture LLM distillation that formulates the problem as a preference optimization task. Unlike standard CoT distillation, the approach transfers knowledge through diverse reasoning traces. It employs an Odds-Ratio Preference Optimization objective that contrasts teacher and student traces for more effective learning, and adopts a mixed-policy strategy for utilizing student-generated outputs, outperforming both off- and on-policy alternatives. Experiments on five datasets and multiple student models show consistent improvements over conventional black-box KD baselines.
\end{abstract}

\section{Introduction}

Knowledge distillation (KD) has emerged as a key approach for compressing LLMs into smaller, more efficient task-specific student models. While white-box KD techniques depend on shared token vocabularies and teacher logits, limiting them to same-architecture settings, black-box KD enables cross-architecture transfer by sampling teacher outputs for supervision. In this work, we introduce \textit{ORPO-Distill}, a general-purpose technique for cross-architecture LLM distillation that reformulates the process as a preference optimization problem. ORPO-Distill integrates three key insights: 
\begin{enumerate}
 \item Distilling from \textit{diverse} reasoning traces improves supervision over single CoT distillation.
\item Framing distillation as a \textit{preference optimization} task using ORPO objective, where teacher-generated positive reasoning and student-generated negative reasoning strengthens contrastive learning.
\item \textit{Mixed-policy} update of student negative traces outperforms both off- and on-policy strategies.
\end{enumerate}

We evaluate ORPO-Distill across five QA benchmarks and multiple student architectures, emphasizing each component’s contribution to the pipeline and overall improvements over black-box KD baselines. 

\section{Methodology}

\textbf{Related Work.} \hspace{0.1cm}\citet{hsieh2023distilling} proposed augmenting teacher CoT reasoning traces with labels to provide additional supervision in a multi-task learning fashion. Recent black-box KD methods such as MAGDI \citet{chen2024magdi} and \citet{payoungkhamdee2024empirical} proposed utilizing contrastive learning with diverse positive and negative rationales generated from single or multiple teacher LLMs, but do not use student-generated outputs (SGOs) in their formulation. Prior literature \citet{agarwal2024policy}, \citet{ko2024distillm}, \citet{ko2025distillm} on white-box KD methods provide growing evidence that utilizing SGOs during distillation provides significant gains. This is because it addresses the distribution mismatch between fixed output sequences seen during training and the auto-regressive generation by student at inference. These works further argue that updating SGOs during training through on- or off-policy updates improves performance. \citet{wang2024qcrd} on the other hand, proposes a black-box KD technique that does utilize contrastive negative SGOs, but doesn’t use policy updates from previous iteration student models. 

\begin{figure}[t]
    \centering
    
    \fbox{\includegraphics[width=0.93\linewidth]{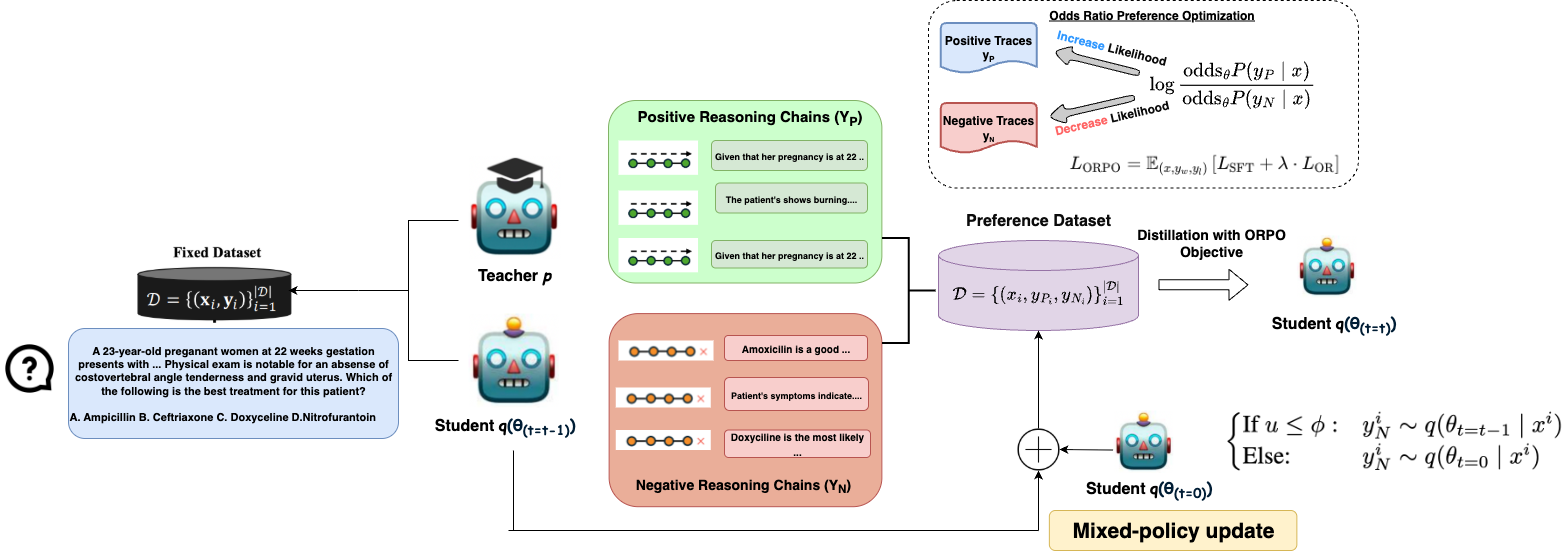}}
    \captionsetup{width=0.95\linewidth}
    \caption{Overview of our ORPO-Distill method. Given an input prompt, teacher and student model generate diverse positive and negative reasoning traces forming the preference dataset for ORPO contrastive distillation, which is updated in a mixed-policy fashion over training epochs.}
    \label{fig:orpo_framework}
\end{figure}

\textbf{Novelty.}\hspace{0.1 cm} In this work therefore, we seek to bridge the gap between these white-box and black-box KD methods through a novel formulation that allows task-specific cross architecture distillation.  In our methodology, we utilize diverse reasoning traces for supervision, combine positive teacher traces and negative student traces through a contrastive ORPO loss objective, and address distribution mismatch between training and auto-regressive inference through a mixed-policy sampling technique for generating negative student reasoning traces. More details about each of these components follows below. A pipeline diagram of our methodology is shown in Figure-\ref{fig:orpo_framework}.

\setlength{\fboxsep}{2pt}   
\setlength{\fboxrule}{0.8pt} 

\subsection{Odds Ratio Preference Optimization (ORPO)}
\label{sec:orpo}

ORPO, proposed by \citet{hong2024orpo}, seeks to combine the traditional two-stage LLM training pipeline comprising of SFT followed by preference alignment stage into a single objective function. This is done by incorporating an odds ratio-based penalty to the conventional negative log-likelihood (NLL) term. This additional penalty term weighted by hyperparameter $\lambda$ imposes a preference for one response over the other. Assuming $y_P$ and $y_N$ denote favoured positive trace and disfavoured negative trace respectively and $q_\theta$ denotes parametrized student model distribution, the objective term for an input sequence $x$ would be defined as:
\begin{equation}
L_{SFT} = - \log q_\theta(y_P \mid x), 
\quad L_{OR} = - \log \sigma \Bigg( \log \frac{ \text{odds}\hspace{0.05cm}q_\theta (y_P|x) }{ \text{odds}\hspace{0.05cm}q_\theta (y_N|x) } \Bigg)
\end{equation}
where, the odds term and the final ORPO objective is defined as:
\begin{equation}
\text{odds}\hspace{0.05cm}q_\theta (y|x) = \frac{ q_\theta(y|x) }{ 1 - q_\theta(y|x) }, \quad L_{ORPO} = L_{SFT} + \lambda L_{OR}
\end{equation}

The authors of ORPO conclude that a small value $\lambda=0.1$ is suited for slightly nudging model towards human preferences(say, helpfulness) whereas a large value such as $\lambda=1$ is used for strong adaptation of one reasoning over other. Since this aligns with our goal of clipping incorrect generation paths in the student model reasoning, we set $\lambda=1$ in our pipeline. 

\subsection{Preference Dataset Creation}

\paragraph{Dataset Structure.} ORPO relies on creating a preference dataset consisting of $\langle \text{Prompt, Chosen, Rejected} \rangle$ triplets. We adopt the same structure, where \textit{Chosen} is a teacher CoT trace leading to the positive or correct answer and \textit{Rejected} is a student CoT trace leading to a negative or incorrect answer. Compared to a previous work \citet{payoungkhamdee2024empirical}, which utilizes teacher CoT traces for creating both positive and negative traces for contrastive training, we first empirically conclude in (Table~\ref{tab:results}) on a subset of datasets 
\begin{wraptable}{r}{0.55\textwidth}
\centering
\small
\captionsetup{justification=centering}
\caption{Effect of sampling contrastive negative trace from Teacher vs Student model (Accuracy \%)}
\label{tab:results}
\begin{tabular}{ccc}
\toprule
\textbf{Experiment} & \multicolumn{2}{c}{\textbf{Datasets}} \\
\cmidrule(lr){2-3}
Off-Policy ORPO & MedQA & ARC-C \\
\midrule
$(p\text{-}CoT_{Teacher},\; n\text{-}CoT_{Teacher})$ & 41.72 & 45.87 \\
    $(p\text{-}CoT_{Teacher},\; n\text{-}CoT_{Student})$ & 49.33 & 56.48 \\
\bottomrule
\end{tabular}
\vspace{-5pt}
\end{wraptable}
that utilizing negative traces from student-generated outputs, infact yields better results for contrastive training using ORPO and thus forms the basis of our further methodology. For this work, we use multi-choice QA datasets with ground truth labels, enabling direct classification of traces as positive or negative. However, the approach generalizes to open-ended tasks given a task-specific definition of desired responses. Such definitions may rely on training small verifier or reward models, unit tests, or other heuristics, and represent an avenue for future work.

\begin{algorithm}[tb] 

\caption{ORPO-Distill}
\label{alg:gkd-orpo}
\begin{algorithmic}[1]
\State \textbf{Given:} Teacher model $p$, Student model $q_\theta$, Fixed dataset $(X, L)$ denoting (prompt,label) pair
\State \textbf{Hyperparameters:} Policy fraction $\phi \in [0,1]$, Diversity Param $K$, Odds-Ratio $\lambda$, LR $\eta$ 
\State \textbf{Initialize:} Sample K diverse positive traces $\{y_P^{(k)}\}_{k=1}^{K} \sim p (\cdot \mid x)$ $\forall$ x $\in$ X to get $Y_P$

\For{epoch $e = 1, 2..\dots, E$}
    \For{iteration $t = 1, 2..\dots, T$}
    \State Generate a random number $u \sim \mathcal{U}(0,1)$
        \If{$u \leq \phi$}
   \State Sample $K$ negative traces $\{y_n^{(k)}\}_{k=1}^{K} \sim q(\theta_{t=t-1} \mid x) \rightarrow \mathcal{B} = \{ (x^{i}, y_{P}^{i}, y_{N}^{i}) \}_{i=1}^{|\mathcal{B}|}$

    \Else
    \State Sample $K$ negative traces $\{y_n^{(k)}\}_{k=1}^{K} \sim q(\theta_{t=0} \mid x) \rightarrow \mathcal{B} = \{ (x^{i}, y_{P}^{i}, y_{N}^{i}) \}_{i=1}^{|\mathcal{B}|}$
    \EndIf
    \State Update $\theta$ to minimize $L_{\text{ORPO}}$: $L_{ORPO} = L_{SFT} + \lambda L_{OR}$
\EndFor
\EndFor
\end{algorithmic}
\end{algorithm}

\textbf{Diversity Sampling.} \hspace{0.1 cm} We sample diverse reasoning chains for both positive and negative traces using temperature sampling with parameter $\tau = 0.8$ for $K$ generations. Both teacher and student are prompted with the same \texttt{Reason-then-Answer} format, where the final answer is marked in a parseable boxed \{\} output. Note, that we do not induce any bias in the prompt while sampling negative traces such as by injecting the incorrect answer. Experiments with $K \in \{4, 8, 12\}$ showed diminishing gains beyond $8$, so we fix $K=8$ in the following experiments. To remove redundancy, we apply rejection sampling and discard traces with high ROUGE-L overlap over $0.80$.

\subsection{Mixed-Policy Update}
We define updates to student-generated negative traces under three settings: off-policy - which corresponds to a fixed negative set generated using initial student model, on-policy - which corresponds to a new negative trace set sampled after every epoch using the latest checkpoint, and mixed-policy - which corresponds to using a combination of last checkpoint and base model traces. The pseudocode of our method is detailed in Algorithm-\ref{alg:gkd-orpo} covering all these cases using a policy fraction parameter $\phi$.

\section{Experimentation}

\textbf{Datasets} We conduct extensive experiments on five widely-used benchmark QA datasets across different domains such as: MedQA-USMLE for medical diagnostic reasoning, ARC-Challenge, StrategyQA and OpenBookQA for general-purpose reasoning, and GSM8K for mathematical problem-solving reasoning. 

\textbf{Implementation Details} We use \texttt{InternLM 2.5 7B-Chat} as the teacher model, selected for its strong performance and instruction-following ability with reliable adherence to the \texttt{Reason-then-Answer} format in zero-shot prompting. However, as discussed, our method is invariant to any teacher and student architecture combinations. We use two student models from different architecture families and sizes: \texttt{InternLM 2.5 1.8B-Chat}, \texttt{TinyLlama 1.1B-Instruct} to validate our method. We conduct full-parameter tuning for 5 epochs, with additional implementation and training details available in our codebase.

\textbf{Experiments} We design experiments to validate each component of our methodology as below, and report results across all datasets in Table~\ref{experimental-results}.

\begin{enumerate}
    \item \textbf{Diverse Reasoning Traces.} We validate this component by creating a single-trace CoT fine-tuning baseline against diverse-traces CoT fine-tuning using the same NLL objective. Both settings use traces generated from the teacher model. The latter equates to SeqKD, proposed by \citet{kim2016sequence}, in the SFT setting and has been used as a baseline in previous literature \citet{payoungkhamdee2024empirical}, 
    and \citet{ko2024distillm}. We also show baseline results of zero-shot CoT evaluation to measure the models out-of-box performance. 
    
    \item \textbf{Contrastive Training with ORPO.} To validate this component, we treat the previous best experiment, i.e., diverse CoT fine-tuning from teacher traces, as the baseline, and seek to improve it further by utilizing the student model generated negative traces using a contrastive ORPO objective. Note that we utilize the base student model (before distillation) at Epoch 0 to generate these negative traces without further updates, and define this as the \textit{Off-Policy ORPO distillation} experiment in results. 
    
    \item \textbf{Mixed-Policy Updates.} The policy parameter $\phi$ defined in our pseudocode Algorithm-\ref{alg:gkd-orpo} controls whether the method is off-policy($\phi=0$), on-policy($\phi=1$), or mixed-policy($\phi=0.5$). This parameter is responsible for controlling the level of mixing between the base student model and the latest epoch checkpoint. For our mixed-policy experiments, we set $\phi = 0.5$ inspired from a similar setting in \citet{agarwal2024policy}, which amounts to an equally proportioned mixing between the two distributions.
\end{enumerate}

\begin{table}
  \caption{Experimental results across different student models and datasets.}
  \label{experimental-results}
  \centering
  \resizebox{0.95\textwidth}{!}{%
  \begin{tabular}{lcccccc}
    \toprule
    \textbf{Experiments} & \multicolumn{5}{c}{\textbf{Datasets (Accuracy \%)}} & \textbf{Avg Acc\%} \\
    \cmidrule(lr){2-6}
    & MedQA & ARC-C & StrategyQA & OBQA & GSM8K & \\
    \midrule
    \multicolumn{7}{l}{\textbf{TinyLlama 1.1B-Instruct}} \\
    \quad Zero-shot CoT Eval & 29.78 & 29.95 & 43.52 & 26.60 & 11.97 & 28.36 \\
    \quad Single CoT Fine Tuning & 32.10 & 32.63 & 46.25 & 29.05 & 31.56 & 34.32 \\
    \quad Diverse CoT Fine Tuning & 34.85 & 35.40 & 47.84 & 33.60 & 36.22 & 37.58 \\
    \quad Off Policy ORPO & 38.95 & 41.20 & 49.77 & 37.45 & 39.45 & 41.36 \\
    \quad On Policy ORPO & 35.11 & 38.01 & 49.24 & 35.60 & 36.88 & 38.97 \\
    \quad Mixed Policy ORPO & \textbf{40.25} & \textbf{43.55} & \textbf{51.25} & \textbf{40.10} & \textbf{40.72} & \textbf{43.17} \\
    \midrule
    \multicolumn{7}{l}{\textbf{InternLM 2.5 1.8B-Chat}} \\
    \quad Zero-shot CoT Eval & 35.82 & 37.12 & 54.15 & 27.40 & 41.02 & 39.10 \\
    \quad Single CoT Fine Tuning & 37.94 & 40.45 & 57.50 & 41.35 & 44.38 & 44.32 \\
    \quad Diverse CoT Fine Tuning & 40.56 & 42.15 & 58.66 & 54.50 & 47.50 & 48.67 \\
    \quad Off Policy ORPO & 49.33 & 56.48 & 59.39 & 53.20 & 51.25 & 53.93 \\
    \quad On Policy ORPO & 43.25 & 49.80 & 58.50 & 52.79 & 47.94 & 50.46 \\
    \quad Mixed Policy ORPO & \textbf{50.43} & \textbf{59.32} & \textbf{61.75} & \textbf{55.22} & \textbf{52.47} & \textbf{55.84} \\
    \midrule
    \multicolumn{7}{l}{\textbf{InternLM 2.5 7B-Chat Teacher}} \\
    \quad Zero-shot CoT Eval Teacher & 50.98 & 56.40 & 62.57 & 61.80 & 66.14 & 59.58 \\
    \bottomrule
  \end{tabular}
    }
\end{table}

\section{Conclusion}
Experiments across multiple student models and datasets show that ORPO-Distill, which leverages student-generated negative traces for contrastive distillation with mixed-policy updates, achieves the best performance. In contrast, purely on-policy updates after every epoch degrade performance compared to the fixed-trace off-policy setting which aligns with findings from \citet{ko2024distillm}. We attribute this to the fact, that although recently sampled negative traces are of higher quality and closely resemble correct rationales, the overall distribution narrows, reducing diversity for contrastive learning. Mixed-policy updates mitigate this issue by anchoring the negative trace distribution to the initial student model (Epoch 0) through random sampling between the initial and most recent checkpoint. We foresee that sophisticated strategies such as curriculum based updates to the mixed-policy buffer and extension to open-ended tasks beyond QA remains an avenue for future work.




\bibliography{reference}
\bibliographystyle{unsrtnat}











\end{document}